\documentclass[letterpaper, 10 pt, conference]{ieeeconf}  

\IEEEoverridecommandlockouts                              
\usepackage[letterpaper, left=0.75in, right=0.75in, bottom=0.75in, top=0.75in]{geometry}

\usepackage{url}
\usepackage{caption}
\usepackage{subcaption}
\usepackage{graphicx}
\usepackage{cite}
\usepackage{multirow}
\usepackage{enumerate}
\usepackage{amsmath}
\usepackage{etoolbox}
\usepackage{placeins}
\usepackage{hyperref}
\usepackage{setspace} 
\usepackage{romannum}
\usepackage{wasysym}

\usepackage{bbding}
\usepackage{pifont}
\usepackage{amssymb}
\usepackage{xcolor}
\usepackage[outline]{contour}
\DeclareMathAlphabet\mathbfcal{OMS}{cmsy}{b}{n}

\title{\LARGE \bf Unseen Object Amodal Instance Segmentation \\via Hierarchical Occlusion Modeling}

\author{Seunghyeok Back, Joosoon Lee, Taewon Kim, Sangjun Noh, Raeyoung Kang, Seongho Bak, Kyoobin Lee†
\thanks{All authors are with the School of Integrated Technology (SIT), Gwangju Institute of Science and Technology (GIST), Cheomdan-gwagiro 123, Buk-gu, Gwangju 61005, Republic of Korea. 
† Corresponding author: Kyoobin Lee {\tt\small kyoobinlee@gist.ac.kr}}%
}

\begin{document}

\maketitle
\thispagestyle{empty}
\pagestyle{empty}

\begin{abstract}

Instance-aware segmentation of unseen objects is essential for a robotic system in an unstructured environment. Although previous works achieved encouraging results, they were limited to segmenting the only visible regions of unseen objects. For robotic manipulation in a cluttered scene, amodal perception is required to handle the occluded objects behind others. This paper addresses Unseen Object Amodal Instance Segmentation (UOAIS) to detect 1) visible masks, 2) amodal masks, and 3) occlusions on unseen object instances. For this, we propose a Hierarchical Occlusion Modeling (HOM) scheme designed to reason about the occlusion by assigning a hierarchy to a feature fusion and prediction order. We evaluated our method on three benchmarks (tabletop, indoors, and bin environments) and achieved state-of-the-art (SOTA) performance. Robot demos for picking up occluded objects, codes, and datasets are available at \url{https://sites.google.com/view/uoais}.
\end{abstract}

\section{INTRODUCTION}

Segmentation of unseen objects is an essential skill for robotic manipulations in an unstructured environment. Recently, unseen object instance segmentation (UOIS) \cite{danielczuk2019segmenting,xie2020best,xie2021unseen,xiang2020learning, durner2021unknown, back2020segmenting} have been proposed to detect unseen objects via category-agnostic instance segmentation by learning a concept of object-ness from large-scale synthetic data. However, these methods focus on perceiving only visible regions, while humans have the ability to infer the entire structure of occluded objects based on the visible structure \cite{palmer1999vision, zhu2017semantic}. This capability, called amodal perception, can allow a robot to straightforwardly manipulate the occlusions in a cluttered scene. Although recent works \cite{wada2019joint, wada2018instance, inagaki2019detecting, price2019inferring, narasimhan2020seeing, qin2020s4g, li2021robotic} have shown the usefulness of amodal perception in robotics, they have been limited to perceiving bounded object sets, where prior knowledge about the manipulating target is given (e.g., labeling a task-specific dataset and training for specific objects and environments).

\begin{figure}[ht!]
    \centering
        \begin{subfigure}[t]{\columnwidth}
            \centering
            \includegraphics[width=\textwidth]{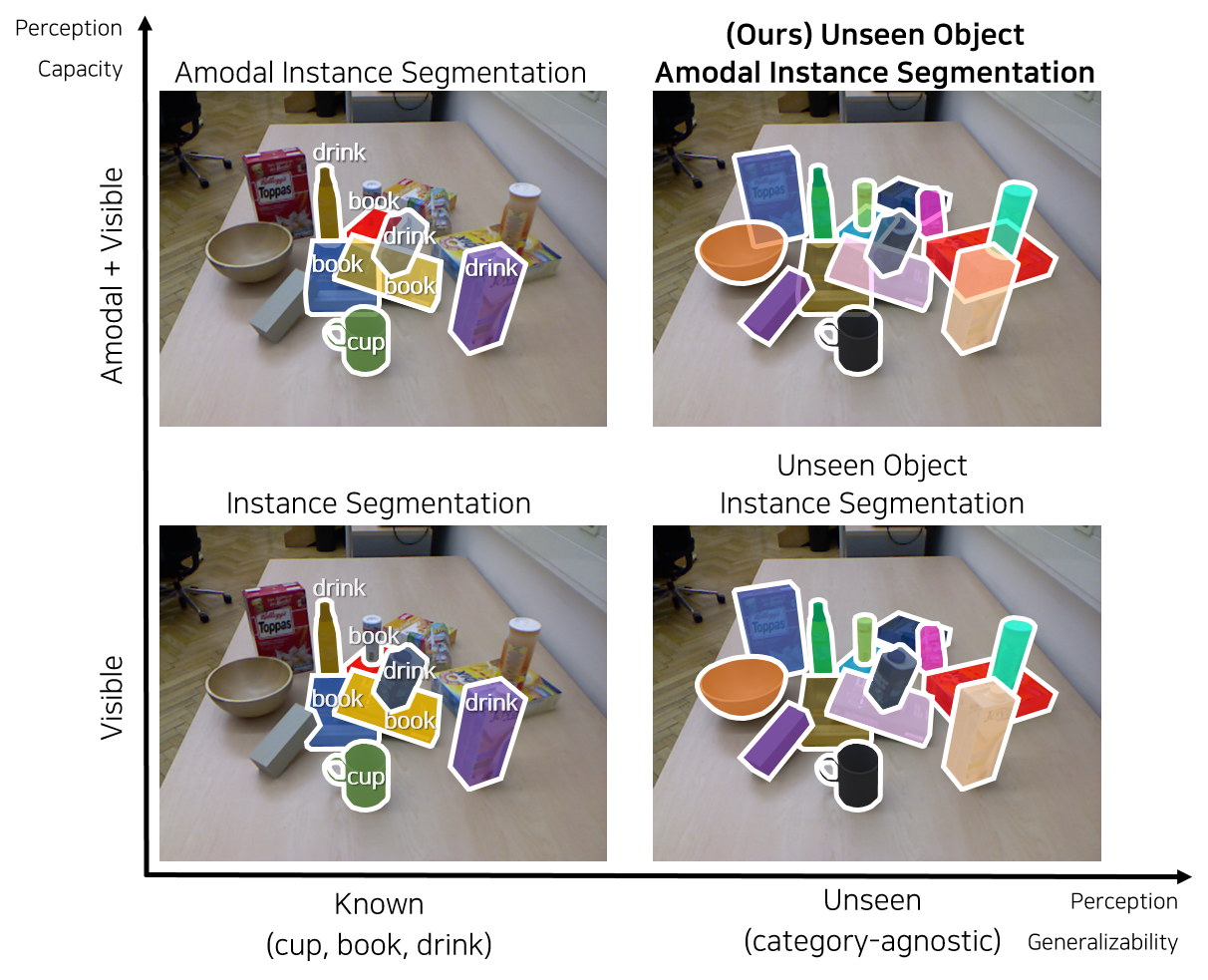}
        \end{subfigure}
        
    \caption{\textbf{Comparison of UOAIS and existing tasks}. While other segmentation methods are limited to perceiving known object sets or detecting only visible masks of unseen objects, UOAIS aims to segment both visible and amodal masks of unseen object instances in unstructured clutter.}
    \label{fig:exemplary_objects}
\end{figure}

This work proposes \textit{unseen object amodal instance segmentation} (UOAIS) to detect visible masks, amodal masks, and occlusion of unseen object instances (Fig. 1). Similar to UOIS, it performs category-agnostic instance segmentation to distinguish the visible regions of unseen objects. Meanwhile, UOAIS jointly performs two additional tasks: amodal segmentation and occlusion classification of the detected object instances. For this, we propose UOAIS-Net, which reasons the object's occlusion via hierarchical occlusion modeling (HOM). The hierarchical fusion (HF) module in our model combines the multiple features of prediction heads according to their hierarchy, thereby allowing the model to consider the relationship between the visible mask, amodal mask, and occlusion. We trained the model on 50,000 photo-realistic RGB-D images to learn various object geometry and occlusion scenes and evaluated its performance in various environments. The experiments demonstrated that visible masks, amodal masks, and the occlusion of unseen objects could be detected in a single framework with state-of-the-art (SOTA) performance. The ablation studies demonstrated the effectiveness of the proposed HOM for UOAIS.

The contributions of this work are summarized as follows:
\begin{itemize}
    \item We propose a new task, UOAIS, to detect category-agnostic visible masks, amodal masks, and occlusion of arbitrary object instances in a cluttered environment.
    \item We propose a HOM scheme to reason about the occlusion of objects by assigning the hierarchy to feature fusion and prediction order.
    \item We introduce a large-scale photorealistic synthetic dataset named UOAIS-SIM and amodal annotations for the existing UOIS benchmark, OSD \cite{richtsfeld2012segmentation}.
    \item We validated our UOAIS-Net on three benchmarks and showed the effectiveness of HOM by achieving state-of-the-art performance in both UOAIS and UOIS tasks.
    \item We demonstrated a robotic application of UOAIS. Using UOAIS-Net, the object grasping order for retrieving the occluded objects in clutter can be easily planned.
\end{itemize}

\section{Related Work}

\noindent\textbf{Unseen Object Instance Segmentation.} UOIS aims to segment the visible regions of arbitrary object instances in an image \cite{xie2020best, xie2021unseen}, and is useful for robotic tasks such as grasping \cite{sundermeyer2021contact} and manipulating unseen objects \cite{murali20206}. Many segmentation methods \cite{felzenszwalb2004efficient, rusu2010semantic, richtsfeld2012segmentation, koo2014unsupervised, potapova2014incremental, pham2018scenecut} have been proposed to distinguish objects but segmentation in cluttered scenes is challenging, especially for objects with complex textures or that are under occlusion \cite{suchi2019easylabel, pham2018scenecut}. To generalize over unseen objects and clutter scenes, recent UOIS methods \cite{danielczuk2019segmenting, xie2020best, xie2021unseen, xiang2020learning,back2020segmenting,durner2021unknown} have trained category-agnostic instance segmentation models to learn the concept of object-ness from a large amount of domain-randomized \cite{tobin2017domain} synthetic data. Although these methods are promising, they focus on segmenting only the visible part of objects. On the other hand, our method can jointly perform visible segmentation, amodal segmentation, and occlusion classification for unseen object instances.

\noindent\textbf{Amodal Instance Segmentation.} When humans perceive an occluded object, they can guess the entire structure even though part of it is invisible \cite{palmer1999vision, zhu2017semantic}. To mimic this amodal perception ability, amodal instance segmentation \cite{li2016amodal} has been proposed, in which the goal is to segment both the amodal and visible masks of each object instance in an image. The SOTA approaches are mainly built on visible instance segmentation \cite{he2017mask, tian2019fcos} and perform amodal segmentation through the addition of an amodal module, such as amodal branch and invisible mask loss \cite{follmann2019learning}, multi-level coding (MLC) \cite{qi2019amodal}, refinement layers \cite{xiao2020amodal}, and occluder segmentation \cite{ke2021deep}. They have demonstrated that it is possible to segment the amodal masks of occluded objects on various datasets \cite{qi2019amodal, follmann2019learning}. However, these methods can detect only a particular set of trained objects and require additional training data to deal with unseen objects. In contrast, UOAIS learns to segment the category-agnostic amodal mask, reducing the need for task-specific datasets and model re-training.

\noindent\textbf{Amodal Perception in Robotics.} The amodal concept is useful for occlusion handling and recent studies have utilized amodal instance segmentation for robotic picking systems \cite{wada2018instance,wada2019joint,inagaki2019detecting} to decide the proper picking order for target object retrieval. Amodal perception has also been applied to various robotics tasks including object search \cite{price2019inferring, danielczuk2020x}, grasping \cite{qin2020s4g, wada2018instance, wada2019joint, inagaki2019detecting}, and active perception \cite{yang2019embodied, li2021robotic}, but it is often limited to perceiving the amodality of specific object sets. The works most related to our method are \cite{price2019inferring} and \cite{agnew2020amodal}. These studies trained a 3D shape completion network that infers the occluded geometry of unseen objects based on visible observation for robotics manipulation in a cluttered scene. Although the reconstruction of the occluded 3D structure is valuable, their amodal reconstruction network requires an additional object instance segmentation module \cite{pham2018scenecut, xie2020best} and only a single instance can be reconstructed on a single forward pass. Whereas their methods require a high computational cost, our method can directly predict the occluded regions of multiple object instances in a single forward pass; thus, it can be easily extended to various amodal robotic manipulations in the real world.

\section{Problem Statement}

We formalize our problem using the following definitions.
\begin{enumerate}[I.]
    \item \textit{Scene}: Let $\mathcal{S}=\{\mathcal{F}_1, ..., \mathcal{F}_K, \mathcal{G}_1, ..., \mathcal{G}_L, \mathcal{E}\}$ be a simulated or real scene with $K$ foreground object instances $\mathcal{F}$, $L$ background object instances $\mathcal{G}$, and a camera $\mathcal{E}$. 
    
    \item \textit{State}: Let $\mathbf{y} =\{\mathcal{B}_k, \mathcal{V}_k, \mathcal{A}_k, \mathcal{O}_k, \mathcal{C}_k\}^K_1$ be a ground truth state of ${K}$ foreground object instances in the scene $S$, which is the set of the bounding box $\mathcal{B}\in \mathbb{R}^{4}$, visible mask $\mathcal{V}\in \{0, 1\}^{W\times H}$, amodal mask $\mathcal{A}\in \{0, 1\}^{W\times H}$, occlusion $\mathcal{O}\in \{0, 1\}$, and class $\mathcal{C}\in \{0, 1\}$ captured by the camera $\mathcal{E}$. $W$ and $H$ are the width and height of the image. The occlusion $\mathcal{O}$ and class $\mathcal{C}$ denote whether the instance is occluded and whether the instance is a foreground objects or not, respectively. 

    \item \textit{Observation}: An RGB-D image $\mathcal{I} = \{\mathcal{R}, \mathcal{U}\}$ captured the scene $\mathcal{S}$ with the camera $\mathcal{E}$ at the pose $\mathcal{T}$. $\mathcal{R}\in \mathbb{R}^{W\times H \times 3}$ is the RGB image, and $\mathcal{U}\in \mathbb{R}^{W\times H}$ is the depth image.  
    
    \item \textit{Dataset and Object Models}: Let $\mathcal{D}=\{(\mathbf{y}_n, \mathcal{I}_n)\}_1^N$ be a dataset which is the set of $N$ RGB-D images $\mathcal{I}$ and corresponding ground truth states $\mathbf{y}$. Let $\mathcal{M_\mathcal{D}}$ be a set of object models used for $\mathcal{F}$ and $\mathcal{G}$ in the dataset $\mathcal{D}$.
    \item \textit{Known and Unseen Object}: Let the train and test set be $\mathcal{D}_{train}$ and $\mathcal{D}_{test}$. Let $f$ be a function trained on $\mathcal{D}_{train}$ and then tested on $\mathcal{D}_{test}$. If the $\mathcal{M}_{\mathcal{D}_{train}}\cap\mathcal{M}_{\mathcal{D}_{test}} = \emptyset$, the object in $\mathcal{M}_{\mathcal{D}_{test}}$ is the unseen object, and the object in $\mathcal{M}_{\mathcal{D}_{train}}$ is the known object for $f$. 
\end{enumerate}

The objective of \textbf{UOAIS} is to detect a category-agnostic visible mask, an amodal mask and the occlusion of arbitrary objects. Thus, our paper aims to find a function $f: \mathcal{I} \rightarrow \mathbf{y}$ for $\mathcal{D}_{test}$ given $\mathcal{D}_{train}$, where $\mathcal{M}_{\mathcal{D}_{train}}\cap\mathcal{M}_{\mathcal{D}_{test}} = \emptyset$.

   \begin{figure*}[ht!]
   \centering
      \includegraphics[width=\textwidth]{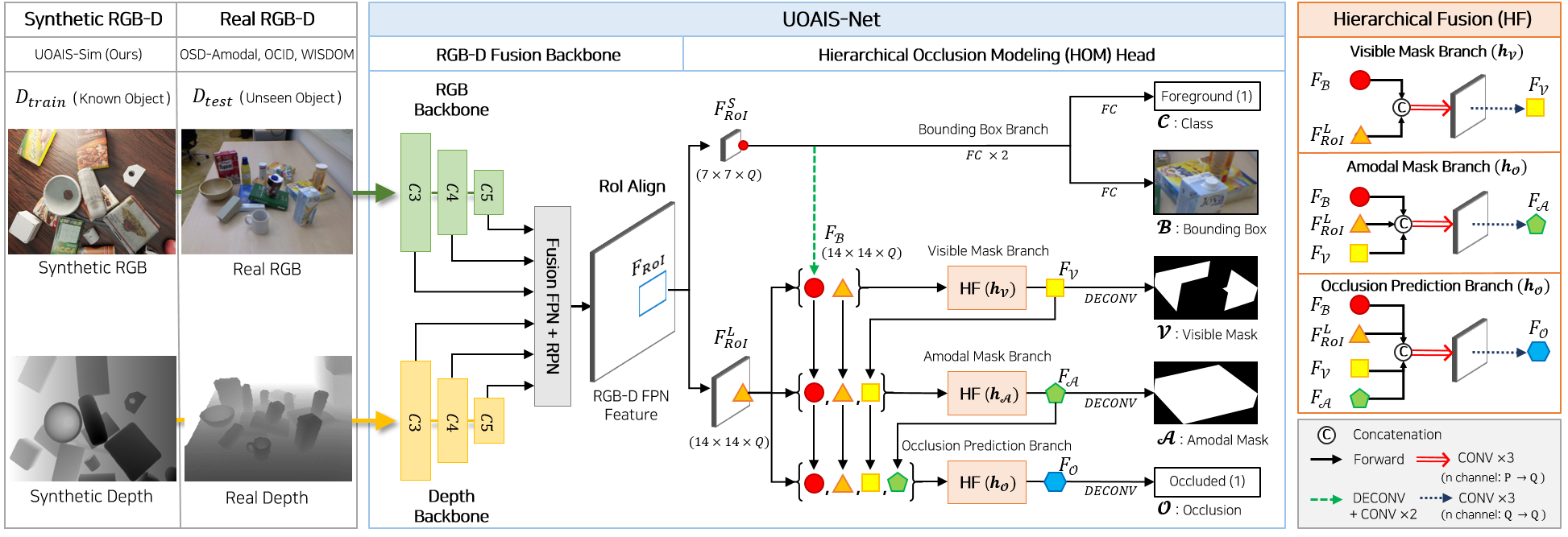}
      \caption{\textbf{Architecture of our proposed UOAIS-Net.} UOAIS-Net consists of (1) an RGB-D fusion backbone for RoI RGB-D feature extraction and (2) an HOM Head for hierarchical prediction of the bounding box, visible mask, amodal mask, and occlusion. It was trained on synthetic RGB-D images and then tested on real clutter scenes. Through occlusion modeling with a HF module, segmentation and occlusion prediction of unseen objects can be significantly enhanced.}
      \label{fig:model}
    \end{figure*}

\section{Unseen Object Amodal Instance Segmentation}

    \subsection{Motivation} \label{motivation}
    To design an architecture for UOAIS, we first made the following observations based on the relationship of bounding box $\mathcal{B}$, visible mask $\mathcal{V}$, amodal mask $\mathcal{A}$, and occlusion $\mathcal{O}$.
    \begin{enumerate}
        \item $\mathcal{B} \rightarrow \mathcal{V, A, O}$: Most SOTA instance segmentation methods \cite{ren2015faster, he2017mask, lee2020centermask, qi2019amodal, chen2019hybrid, tian2019fcos} have detect-then-segment approaches; They detect the object region of interest (RoI) bounding box first, then segment the instance mask in a given RoI bounding box. We followed this paradigm by adopting a Mask R-CNN \cite{he2017mask} for UOAIS, and thus visible, amodal, and occlusion predictions strongly depend on the bounding box prediction.
        \item $\mathcal{V} \rightarrow \mathcal{A}$: The amodal mask is the union of the visible mask $\mathcal{V}$ and the invisible mask $\mathcal{IV}$ ($\mathcal{A}=\mathcal{V}\cup \mathcal{IV}$). The visible mask is more obvious than amodal and invisible masks; thus, segment the visible mask, and then infer the amodal mask based on the segmented visible mask. 
        \item $\mathcal{A,V} \rightarrow \mathcal{O}$: The occlusion is defined by the ratio of the visible mask to the amodal mask, i.e., if the visible mask equals to the amodal mask ($\mathcal{V}/\mathcal{A}=1$), the object is not occluded ($\mathcal{O}=0$); segment the visible and amodal masks, and then classify the occlusion.
    \end{enumerate}
    
    Based on these observations, we propose an \textit{HOM} scheme for UOAIS; (1) detect the RoI bounding box $\mathcal{B}$ (2) segment the visible mask $\mathcal{V}$ (3) segment the amodal mask $\mathcal{A}$, and (4) classify the occlusion $\mathcal{O}$ ($\mathcal{B}\rightarrow\mathcal{V}\rightarrow\mathcal{A}\rightarrow\mathcal{O}$). For this, we propose a UOAIS-Net, which employs an HOM scheme on the top of Mask R-CNN \cite{he2017mask} with an HF module. The HF module explicitly assigns a hierarchy to feature fusion and prediction order and improves the overall performance. 
    
    \subsection{UOAIS-Net}
    \noindent\textbf{Overview.} UOAIS-Net consists of (1) an RGB-D fusion backbone and (2) an HOM Head (Fig. \ref{fig:model}). From an input RGB-D image $\mathcal{I}$, the RGB-D fusion backbone with a feature pyramid network (FPN) \cite{seferbekov2018feature} extracts an RGB-D FPN feature. Next, the region proposal network (RPN) \cite{ren2015faster} proposes possible object regions and the RoIAlign layer crops an RoI RGB-D FPN feature ($F_{RoI}^{S}$) size of $7\times 7\times Q$. Then, the bounding box branch in the HOM Head performs a box regression and foreground classification. For positive RoIs, the HOM Head extracts an RoI RGB-D FPN feature ($F_{RoI}^{L}$) with dimensions of $14\times 14\times Q$ and predicts a visible mask $\mathcal{V}$, amodal mask $\mathcal{A}$, and occlusion $\mathcal{O}$ hierarchically for each RoI via the HF module. $Q$ is the number of channels.
    
    \noindent\textbf{Sim2Real Transfer.} To generalize over unseen objects of various shapes and textures, our model is trained via Sim2Real transfer; trains a model on large synthetic data and then applies it to real clutter scenes. In this scheme, the domain gap between simulation and real scenes greatly affects the model performance \cite{kim2021acceleration, xie2021unseen}. While depth shows a reasonable performance against the Sim2Real transfer \cite{mahler2019learning, danielczuk2019segmenting, back2020segmenting}, training with only the non-photorealistic RGB is often insufficient to segment the real object instances \cite{xie2020best, xie2021unseen}. To address this issue, we trained a UOAIS-Net using photo-realistic synthetic images (UOAIS-Sim in Section \ref{amodal_datasets}) so the Sim2Real gap could be significantly reduced. 

    \noindent\textbf{RGB-D Fusion.} Depth provides useful 3D geometric information for recognizing the object \cite{xie2020best, xie2021unseen, back2020segmenting}, but the joint use of the RGB with depth is required to produce a precise segmentation boundary \cite{hazirbas2016fusenet, park2017rdfnet, xie2020best}, as depth is noisy especially for transparent and reflective surfaces. To effectively learn discriminative RGB-D features, the RGB-D fusion backbone extracts RGB and depth features with a separate ResNet-50 \cite{he2016deep} for each modality. Then, the RGB and depth features are fused into RGB-D features in a multiple level (C3, C4, C5) through concatenation and $1\times1$ convolution, thereby reducing the number of the channel $2Q$ to $Q$. Finally, RGB-D features are fed into the FPN \cite{seferbekov2018feature} and the RoIAlign layer \cite{he2017mask} and form an RGB-D FPN feature. 

    
    \subsection{Hierarchical Occlusion Modeling} 
    \label{HOM}
    The HOM Head is designed to reason about the occlusion in an RoI by predicting the output in the order of $\mathcal{B} \rightarrow \mathcal{V} \rightarrow \mathcal{A} \rightarrow \mathcal{O}$. The HOM Head consists of (1) a bounding box branch (2) a visible mask branch (3) an amodal mask branch, and (4) an occlusion prediction branch. The HF module $h$ for each branch fuses all prior features into that branch.

    \noindent\textbf{Bounding Box Branch ($\mathcal{B}$, $\mathcal{C}$).} The HOM Head first predicts the bounding box $\mathcal{B}$ and the class $\mathcal{C}$ from $F_{RoI}^S$ given by RPN. The structure of the bounding box branch follows a standard localization layer in \cite{he2017mask}. $F_{RoI}^S$ is fed into two fully connected (FC) layers, then the $\mathcal{B}$ and $\mathcal{C}$ are predicted by the following FC layer. For the category-agnostic segmentation, we set $\mathcal{C} \in \{0,1\}$ to detect all the foreground instances.
    
    \noindent\textbf{HF module. ($\mathcal{V}, \mathcal{A}, \mathcal{O}$)} For the positive RoI, the HOM Head performs visible segmentation, amodal segmentation, and occlusion classification sequentially with the HF module. Following motivation $1)$, we first provided a box feature $F_{\mathcal{B}}$ to all other subsequent branches for the prediction conditioned on $\mathcal{B}$. It is similar to the MLC \cite{qi2019amodal} that provides a box feature in mask predictions, which allows the mask layers to exploit global information and enhances the segmentation. $F_{RoI}^S$ is fed into a $3\times 3$ deconvolution layer, and then an upsampled RoI feature with a size of $14\times 14 \times Q$ is forwarded to three $3\times 3$ convolutional layers. The output of this operation is used as the box feature $F_{\mathcal{B}}$.
    
    Following the hierarchy of the HOM scheme, the HF module $h$ for each branch fuses an $F_{RoI}^L$ with $F_{\mathcal{B}}$ and features from a prior branch as follows:
    \begin{equation}
      F_\mathcal{V} = h_\mathcal{V}(F_\mathcal{B}, F_{RoI}^L)
    \end{equation}
    \begin{equation}
      F_\mathcal{A} = h_\mathcal{A}(F_\mathcal{B}, F_{RoI}^L, F_\mathcal{V})
    \end{equation}
    \begin{equation}
      F_\mathcal{O} = h_\mathcal{O}(F_\mathcal{B}, F_{RoI}^L, F_\mathcal{V}, F_\mathcal{A})
    \end{equation}
    where $F_\mathcal{V}$, $F_\mathcal{A}$, and $F_\mathcal{O}$ are the visible, amodal, and occlusion features, respectively, and $h_\mathcal{V}$, $h_\mathcal{A}$, $h_\mathcal{O}$ are the HF modules for the visible, amodal, occlusion branches. Specifically, the HF module fuses all inputs by concatenating them to be the channel dimension as $P$ and passing it to three 3x3 convolutional layers to reduce the number of channels into $Q$. Then it is forwarded to three $3\times 3$ convolutional layers to extract the task-relevant feature in each branch. Finally, each branch of the HOM outputs the final predictions with the prediction layer $g$. The loss for HOM Head $\mathcal{L}_{HOM}$ is 
    \begin{equation}
      {\lambda}_1\mathcal{L}(g_\mathcal{V}(F_\mathcal{V}), \mathcal{V})+{\lambda}_2\mathcal{L}(g_\mathcal{A}(F_\mathcal{A}), \mathcal{A})+{\lambda}_3\mathcal{L}(g_\mathcal{O}(F_\mathcal{O}), \mathcal{O})
    \end{equation}
    where $g_{\mathcal{V}}$, $g_{\mathcal{A}}$, and $g_{\mathcal{O}}$ are the prediction layers for the visible mask, amodal mask, and occlusion, respectively. We used a $2\times2$ deconvolutional layer for $g_{\mathcal{V}}, g_{\mathcal{A}}$, and FC layer for $g_{\mathcal{O}}$. The hyper-parameter weights ${\lambda}_1, {\lambda}_2, {\lambda}_3$ are set to 1. The total loss for UOAIS-Net is $\mathcal{L}_{Regression}+\mathcal{L}_{class}+\mathcal{L}_{HOM}$, where the regression and classification loss are from \cite{he2017mask}. 
    
\begin{figure}[ht]
      \includegraphics[width=\columnwidth]{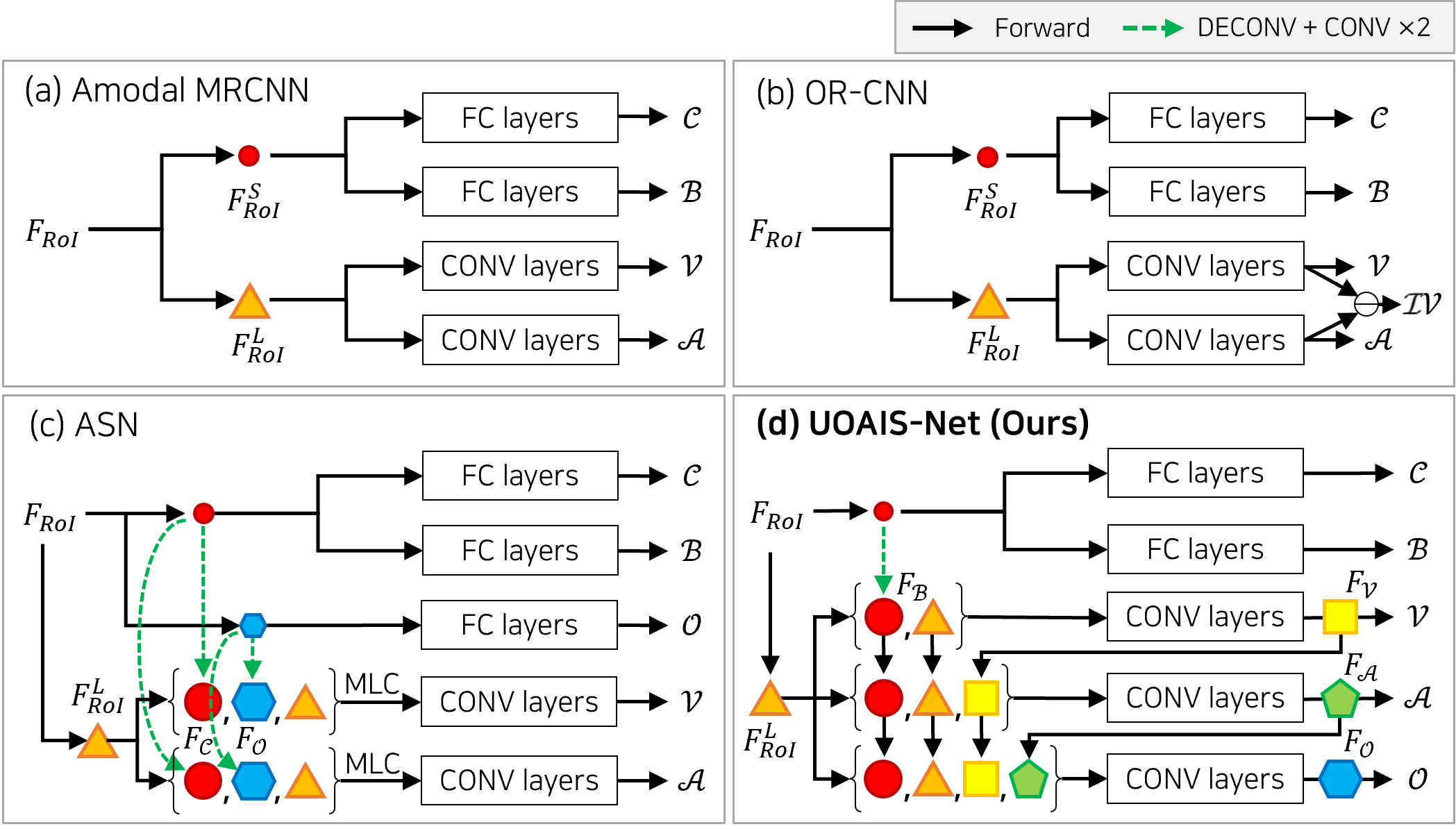}
      \caption{\textbf{Prediction head comparison.} a) Amodal MRCNN \cite{follmann2019learning}, b) ORCNN \cite{follmann2019learning}, c) ASN \cite{qi2019amodal}, d) Ours. UOAIS-Net \textit{hierarchically} fuses the box, visible, amodal and occlusion features ($\mathcal{B} \rightarrow \mathcal{V} \rightarrow \mathcal{A} \rightarrow \mathcal{O}$), while the visible and amodal features in other methods \cite{follmann2019learning, qi2019amodal} are \textit{indirectly} related to each other. $F_{RoI}$ denotes an RoI feature after the RoIAlign.} 
      \label{fig:motivation}
    \end{figure}

    \noindent\textbf{Architecture Comparison.} We compare the prediction heads of amodal instance segmentation methods in Fig. \ref{fig:motivation}. From the cropped RoI feature $F_{RoI}$, Amodal MRCNN \cite{follmann2019learning} outputs all the predictions with separate branches, without considering the relationship between them. ORCNN \cite{follmann2019learning} regulates the invisible mask to be an amodal minus the visible mask, but their features are not inter-weaved. ASN \cite{qi2019amodal} fuses the features from $F_{RoI}^S$ into mask branches via multi-level-coding, but the visible and amodal features are extracted parallelly. In contrast, all the branches in UOAIS-Net are hierarchically connected via the HF module for the hierarchical prediction via occlusion reasoning. 
    
    \begin{figure}[ht!]
        \centering
        \includegraphics[width=\columnwidth]{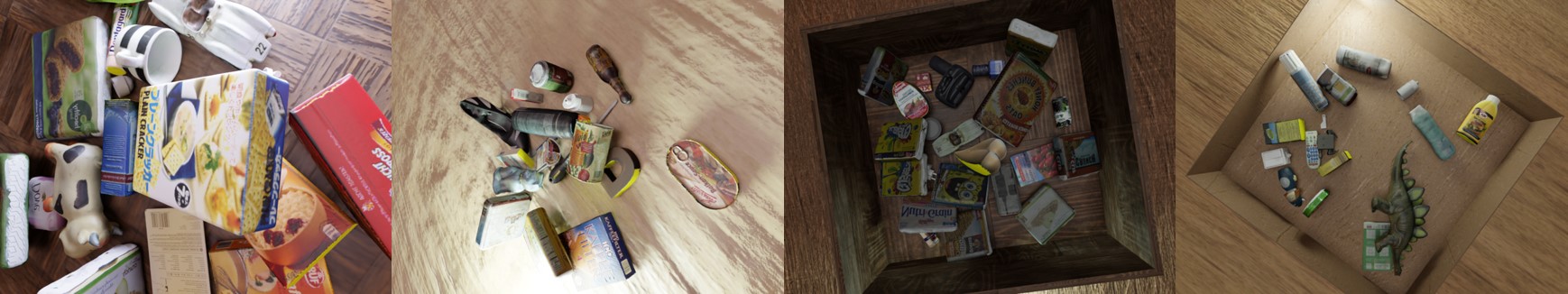}
        \caption{Photo-realistic synthetic RGB images of \textbf{UOAIS-Sim}.}
        \label{fig:uoais-sim}
    \end{figure}
    
    
    \noindent\textbf{Implementation Details.} The model was trained on UOAIS-Sim dataset following the standard schedule in \cite{wu2019detectron2} for $90,000$ iterations with SGD \cite{zinkevich2010parallelized} using the learning rate of $0.00125$. We applied color \cite{liu2016ssd}, depth \cite{zakharov2018keep}, and  crop augmentation. Training took about $8$h on a single Tesla A100, and the inference took $0.13$ s per image (1,000 iterations) on a Titan XP. For it to serve as a general object instance detector, only the plane and bin were set to background objects $\mathcal{G}$ and all other objects are set to foreground objects $\mathcal{F}$; thus, it detected all instances in the image except for the plane and bin. For the cases requiring the task-specific foreground object selection \cite{richtsfeld2012segmentation}, we trained a binary segmentation model \cite{wu2020cgnet} on the TOD dataset \cite{xie2020best}, thereby enabling real-time foreground segmentation ($14$ ms for single forward pass) with less than 0.5 M parameters. 

    \begin{table*}[ht!]
    \caption{\textbf{UOAIS} ($\mathcal{A}$, $\mathcal{IV}$, $\mathcal{O}$, $\mathcal{V}$) performances on \textbf{OSD}\cite{richtsfeld2012segmentation}\textbf{-Amodal}. All methods are trained with RGB-D UOAIS-Sim. $\{\}$ denotes that they are predicted parallelly. $\rightarrow$ refers the hierarchy in prediction heads. OV: Overlap $F$, BO: Boundary $F$}
    \resizebox{\textwidth}{!}{
        \begin{tabular}{|c|c|ccc|ccc|cc|ccc|}
        \hline
        \multirow{2}{*}{Method} & \multirow{2}{*}{Hierarchy Order} & \multicolumn{3}{c|}{Amodal Mask ($\mathcal{A}$)} & \multicolumn{3}{c|}{Invisible Mask ($\mathcal{IV}$)} & \multicolumn{2}{c|}{Occlusion ($\mathcal{O}$)} & \multicolumn{3}{c|}{Visible Mask ($\mathcal{V}$)} \\
        & & OV & BO & $F@.75$ & OV & BO & $F@.75$ & $F_{\mathcal{O}}$ & $ACC_{\mathcal{O}}$ & OV & BO & $F@.75$ \\ \hline
        Amodal MRCNN \cite{follmann2019learning}  & \{ $\mathcal{B} , \ \mathcal{V}, \ \mathcal{A}$ \}  
            & 82.4 & 66.6 & 82.5 & 50.9 & 28.4 & 41.9 & 74.5 & 81.9 & 83.3 & 69.8 & 74.9 \\
        ORCNN \cite{follmann2019learning}   & \{ $\mathcal{B} , \ \mathcal{V} , \  \mathcal{A}$ \} $\rightarrow \mathcal{IV}$ 
            & \textbf{83.1} & 67.2 & 84.1 & 49.2 & 25.8 & 33.6 & 75.5 & 83.1 & 83.2 & 70.1 & 75.9 \\
        ASN \cite{qi2019amodal}             & \{ $\mathcal{B} , \ \mathcal{O}$ \} $ \rightarrow $ \{ $ \mathcal{V} , \ \mathcal{A}$ \}  
            & 80.7 & 67.0 & \textbf{84.5} & 42.0 & 20.7 & 38.0 & 59.1 & 65.7 & 85.1 & 72.9 & 78.8 \\
        \textbf{UOAIS-Net (Ours)}           & $\mathcal{B} \rightarrow \mathcal{V} \rightarrow \mathcal{A} \rightarrow \mathcal{O}$     
            & 82.1 & \textbf{68.7} & 83.7 & \textbf{55.3} & \textbf{32.3} & \textbf{49.2} & \textbf{82.1} & \textbf{90.9} & \textbf{85.2} & \textbf{73.0} & \textbf{79.2} \\ \hline
        \end{tabular}}
        \label{table1}
    \end{table*}
    
    \begin{table*}[ht!]
    \caption{\textbf{UOIS} ($\mathcal{V}$) performances of UOAIS-Net and SOTA UOIS methods on \textbf{OSD} \cite{richtsfeld2012segmentation} and \textbf{OCID} \cite{suchi2019easylabel}.}
    \resizebox{\textwidth}{!}{
        \begin{tabular}{|c|c|c|c|ccc|ccc|c|ccc|ccc|c|}
        \hline
        \multirow{3}{*}{Method} & \multirow{3}{*}{Input} & \multirow{3}{*}{\begin{tabular}[c]{@{}c@{}}Amodal\\ Perception\end{tabular}} & \multirow{3}{*}{\begin{tabular}[c]{@{}c@{}}Refine\\ -ment\end{tabular}} & \multicolumn{7}{c|}{OSD \cite{richtsfeld2012segmentation}} & \multicolumn{7}{c|}{OCID \cite{suchi2019easylabel}} \\ \cline{5-18} 
         &  &  &  & \multicolumn{3}{c|}{Overlap} & \multicolumn{3}{c|}{Boundary} & \multirow{2}{*}{$F@.75$} & \multicolumn{3}{c|}{Overlap} & \multicolumn{3}{c|}{Boundary} & \multirow{2}{*}{$F@.75$} \\
         &  &  &  & P & R & F & P & R & F &  & P & R & F & P & R & F &  \\ \hline
        UOIS-Net-2D \cite{xie2020best} & RGB-D &  &  & 80.7 & 80.5 & 79.9 & 66.0 & 67.1 & 65.6 & 71.9 & 88.3 & 78.9 & 81.7 & 82.0 & 65.9 & 71.4 & 69.1 \\
        UOIS-Net-3D \cite{xie2021unseen} & RGB-D &  &  & \textbf{85.7} & 82.5 & 83.3 & \textbf{75.7} & 68.9 & 71.2 & 73.8 & 86.5 & 86.6 & 86.4 & 80.0 & 73.4 & 76.2 & 77.2 \\
        UCN \cite{xiang2020learning} & RGB-D &  &  & 84.3 & \textbf{88.3} & \textbf{86.2} & 67.5 & 67.5 & 67.1 & 79.3 & 86.0 & \textbf{92.3} & 88.5 & 80.4 & 78.3 & 78.8 & 82.2 \\ 
        Mask R-CNN \cite{he2017mask} & RGB-D &  &  
        & 83.9 & 84.2 & 79.1 & 71.2 & 72.0 & 70.9 & 77.8 & 67.6 & 83.7 & 68.9 & 63.3 & 72.6 & 64.1 & 68.4 \\
        \textbf{UOAIS-Net (Ours)} & RGB & \Checkmark &  
        & 84.2 & 83.7 & 83.8 & 72.2 & 72.8 & 72.1 & 76.7 & 66.5 & 83.1 & 67.9 & 62.1 & 70.2 & 62.3 & 73.1 \\
        \textbf{UOAIS-Net (Ours)} & Depth & \Checkmark &  
        & 84.9 & 86.4 & 85.5 & 68.2 & 66.2 & 66.9 & \textbf{80.8}  & \textbf{89.9} & 90.9 & \textbf{89.8} & \textbf{86.7} & \textbf{84.1} & \textbf{84.7} & \textbf{87.1} \\
        \textbf{UOAIS-Net (Ours)} & RGB-D & \Checkmark &  
        & 85.3 & 85.3 & 85.2 & 72.6 & \textbf{74.2} & \textbf{73.0} & 79.2 & 70.6 & 86.7 & 71.9 & 68.2 & 78.5 & 68.8 & 78.7 \\ \hline
        UCN + Zoom-in \cite{xiang2020learning} & RGB-D &  & \Checkmark & \multicolumn{1}{c}{87.4} & \multicolumn{1}{c}{87.4} & \multicolumn{1}{c|}{87.4} & \multicolumn{1}{c}{69.1} & \multicolumn{1}{c}{70.8} & \multicolumn{1}{c|}{69.4} & \multicolumn{1}{c|}{83.2} & \multicolumn{1}{c}{91.6} & \multicolumn{1}{c}{92.5} & \multicolumn{1}{c|}{91.6} & \multicolumn{1}{c}{86.5} & \multicolumn{1}{c}{87.1} & \multicolumn{1}{c|}{86.1} & \multicolumn{1}{c|}{89.3} \\ \hline
        \end{tabular}
        }
        \label{table2}
    \end{table*}
    
\subsection{Amodal Datasets}
    \label{amodal_datasets}
    \noindent\textbf{UOAIS-Sim.} We generated 50,000 RGB-D images of 1,000 cluttered scenes with amodal annotations (Fig. 4.). We used the BlenderProc \cite{denninger2019blenderproc} for photo-realistic rendering. A total of 375 3D textured object models from \cite{kasper2012kit, singh2014bigbird, hodavn2020bop} were used, including household (e.g., cereal box, bottle) and industrial objects (e.g., bracket, screw) of various geometries. One to 40 objects were randomly dropped on randomly textured bin or plane surfaces. Images were captured at a random camera pose. We split the images and object models with a 9:1 and 4:1 ratio for the training and validation sets. The YCB object models \cite{calli2015ycb} in BOP \cite{hodavn2020bop} were excluded from the simulation so that they could be utilized as test objects in the real world.

    \noindent\textbf{OSD-Amodal.} To benchmark the UOAIS performance, we introduce amodal annotations for the OSD \cite{richtsfeld2012segmentation} dataset. Few amodal instance datasets have been proposed, but \cite{zhu2017semantic} and \cite{qi2019amodal} consist of mainly outdoor scenes, and \cite{follmann2019learning} does not contain depth. OSD is a UOIS benchmark consisting of RGB-D tabletop clutter scenes, but it lacks amodal annotations. Three annotators carefully annotated and cross-checked the amodal instance masks to ensure consistent and precise annotation.

\section{Experiments}
    
\label{section:exp}
    \noindent\textbf{Datasets.} We compared our methods and SOTA on the three benchmarks. OSD\cite{richtsfeld2012segmentation}, consisting of 111 tabletop clutter images, was used to evaluate the UOAIS ($\mathcal{V}$, $\mathcal{A}$, $\mathcal{IV}$, $\mathcal{O}$) and UOIS ($\mathcal{V}$) performances. Further, OCID and WISDOM were used to compare the UOIS performance. OCID provides 2,346 indoor images and WISDOM includes 300 top-view test images in bin picking. The image size of input was set to $640\times 480$ in OSD and OCID, and $512\times 384$ in WISDOM. 
    
    \noindent\textbf{Metrics.} In OSD and OCID, we measured the Overlap $P/R/F$, Boundary $P/R/F$, and $F@.75$ for the amodal, visible, and invisible masks \cite{dave2019towards, xie2020best}. Overlap $P/R/F$ and Boundary $P/R/F$ evaluate the whole area and the sharpness of the prediction, respectively, where $P$, $R$, and $F$ are the precision, recall, and F-measure of instance masks after the Hungarian matching, respectively. $F@.75$ is the percentage of segmented objects with an Overlap F-measure greater than 0.75. For details, refer to \cite{xie2020best} and \cite{dave2019towards}. We also reported the accuracy ($ACC_{o}$) and F-measure ($F_{o}$) of occlusion classification, where $ACC_{o}=\frac{\delta}{\alpha}$, $F_{o}=\frac{2P_{o}R_{o}}{P_{o}+R_{o}}$, $P_{o} =\frac{\delta}{\beta}$, $R_{o}=\frac{\delta}{\gamma}$. $\alpha$ is the number of the matched instances after the Hungarian matching. $\beta$, $\gamma$, and $\delta$ are the numbers of occlusion prediction, ground truth, and correct prediction, respectively. In WISDOM, we measured the mask AP, AP50, and AR \cite{lin2014microsoft}. All the metrics of the model trained on UOAIS-Sim are the average scores of three different random seeds. We used a light-weight network \cite{wu2020cgnet} on the OSD to filter the out-of-the-table instances as described in Section \hyperref[HOM]{\ref{HOM}}. \vspace{0.1cm}

    \begin{figure*}[ht!]
        \centering
        \includegraphics[width=\textwidth]{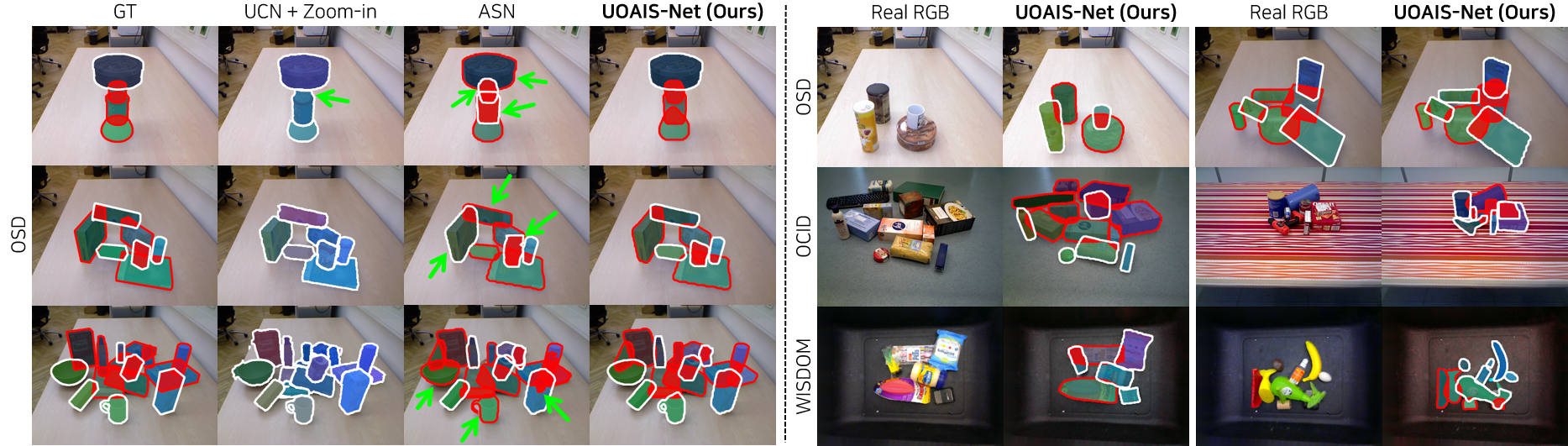}
        \caption{\textbf{Comparison} of UOAIS-Net and SOTA methods (left) and \textbf{predictions} of our methods on various environments (right). Red region: pixel predicted as an invisible mask ($\mathcal{IV}=\mathcal{A}-\mathcal{V}$), Red border: object predicted as an occluded ($\mathcal{O}=1$)}
        \label{fig:uoais-sim-real}
    \end{figure*}

    \noindent\textbf{Comparison with SOTA in UOAIS ($\mathcal{A}$, $\mathcal{IV}$, $\mathcal{O}$, $\mathcal{V}$).} Table \ref{table1} compares the UOAIS-Net and the SOTA amodal instance segmentation methods \cite{follmann2019learning, qi2019amodal} in OSD-Amodal. SOTA methods are trained on UOAIS-Sim RGB-D images with the same hyper-parameters of UOAIS-Net. The only difference between the models is the prediction heads (Fig. \ref{fig:motivation}). The occlusion predictions for Amodal MRCNN and ORCNN are decided using the ratio of $\mathcal{V}$ to $\mathcal{A}$ ($\mathcal{O}=1$ if $\mathcal{V}/\mathcal{A}<0.95$). Our proposed UOAIS-Net outperforms the others on $\mathcal{IV}$, $\mathcal{O}$, and $\mathcal{V}$ and achieves a similar performance with the ASN on $\mathcal{A}$, which shows the effectiveness of the HOM Head.

    \begin{table}[ht!]
    \caption{\textbf{UOIS} ($\mathcal{V}$) performances in WISDOM \cite{danielczuk2019segmenting}}
    \centering
    \resizebox{\columnwidth}{!}{
        \begin{tabular}{|c|c|c|ccc|}
        \hline
        \multirow{2}{*}{Method} & \multirow{2}{*}{Input} & \multirow{2}{*}{Train Dataset} & \multicolumn{3}{c|}{Visible Mask ($\mathcal{V}$)} \\
         &  &  & AP & AP50 & AR \\ \hline
        Mask R-CNN \cite{danielczuk2019segmenting} & RGB & WISDOM-Sim & 38.4 & - & 60.8 \\
        Mask R-CNN \cite{danielczuk2019segmenting} & Depth & WISDOM-Sim & 51.6 & - & 64.7 \\
        Mask R-CNN \cite{danielczuk2019segmenting} & RGB & WISDOM-Real & 40.1 & 76.4 & - \\
        D-SOLO \cite{wang2020solo} & RGB & WISDOM-Real & 42.0 & 75.1 & - \\
        PPIS \cite{ito2020point} & RGB & WISDOM-Real & 52.3 & 82.8 & - \\
        Mask R-CNN & RGB & UOAIS-Sim & 60.6 & 90.3 & 68.8 \\
        Mask R-CNN & Depth & UOAIS-Sim & 56.5 & 85.9 & 65.4 \\
        Mask R-CNN & RGB-D & UOAIS-Sim & 63.8 & 91.8 & 70.5 \\
        \textbf{UOAIS-Net (Ours)} & RGB-D & UOAIS-Sim & \textbf{65.9} & \textbf{93.3} & \textbf{72.2} \\ \hline
        \end{tabular}}
        \label{table3}
    \end{table}

   \noindent\textbf{Comparison with SOTA in UOIS ($\mathcal{V}$).} We compared the visible segmentation of our methods and SOTA methods \cite{xie2020best, xiang2020learning, xie2021unseen, danielczuk2019segmenting, wang2020solo, ito2020point} on OSD and OCID (Table \ref{table2}), and WISDOM (Table \ref{table3}). As the experiment settings are slightly different (i.e., non-photorealistic training images in \cite{xie2020best, xiang2020learning, xie2021unseen}, ResNet-32 \cite{he2016deep} backbone in \cite{danielczuk2019segmenting}), we also show the performance of the Mask R-CNN \cite{he2017mask} trained under the same setting with UOAIS-Net. UOAIS-Net outperformed the Mask R-CNN baselines in all datasets, indicating that amodal segmentation aids the visible segmentation by reasoning over the whole object and explicitly considering occlusion. Compared to SOTA UOIS methods, UOAIS-Net outperformed the others in WISDOM and achieved the performance on par with the other methods in OSD and OCID. Note that our methods can detect amodal masks and occlusions, while UOIS can only segment visible masks. UCN with Zoom-in refinement \cite{xiang2020learning} achieved a better performance than ours in OCID, but it requires more computation ($0.2+0.05K$ s, $K$: number of instances) than ours ($0.13$ s). Also, the loss function of UCN considers only visible regions by pushing the feature embedding from the same objects to be close; thus, an extra algorithm should be added for the amodal perception. The UOAIS-Net with depth performs better than the RGB-D model in OCID, as it includes much more complex textured backgrounds, resulting in foreground segmentation error.

    \begin{table}[htb]
        \centering
        \caption{Ablation of \textbf{hierarchy order} on OSD-Amodal}
        \resizebox{\columnwidth}{!}{
            \centering
            \begin{tabular}{|c|c|ccccccc|}
            \hline
            \multirow{2}{*}{Hierarchy} & \multirow{2}{*}{\begin{tabular}[c|]{@{}c@{}}Overall\\ Score\end{tabular}} & 
            \multicolumn{2}{c}{Amodal Mask} & \multicolumn{2}{c}{Invisible Mask} & Occlusion & \multicolumn{2}{c|}{Visible Mask} \\
            & & OV & $F@.75$ & OV & $F@.75$ & $ACC_{\mathcal{O}}$ & OV & $F@.75$ \\ \hline
            $\mathcal{O} \rightarrow \mathcal{A} \rightarrow \mathcal{V}$ & 0.42 & 80.6 & 84.9 & 53.9 & 49.9 & 89.3 & 85.1 & 79.7 \\
            $\mathcal{A} \rightarrow \mathcal{O} \rightarrow \mathcal{V}$ & 0.44 & 80.8 & 83.7 & 54.3 & 50.7 & 91.3 & 85.3 & 78.7 \\
            $\mathcal{A} \rightarrow \mathcal{V} \rightarrow \mathcal{O}$ & 0.54 & 81.5 & 83.2 & 54.7 & 48.8 & 91.0 & 85.5 & 79.4 \\
            $\mathcal{O} \rightarrow \mathcal{V} \rightarrow \mathcal{A}$ & 0.48 & 81.7 & 84.2 & 55.8 & 51.1 & 89.6 & 84.9 & 78.7 \\
            $\mathcal{V} \rightarrow \mathcal{O} \rightarrow \mathcal{A}$ & 0.50 & 82.4 & 84.1 & 55.4 & 47.9 & 90.3 & 85.2 & 78.8 \\
            $\mathbfcal{V} \rightarrow \mathbfcal{A} \rightarrow \mathbfcal{O}$ & \underline{\textbf{0.57}} & 82.1 & 83.7 & 55.3 & 49.2 & 90.6 & 85.2 & 79.2 \\ \hline
            \end{tabular}}
            \label{table4}
    \end{table}

    \begin{table}[htb]
    \caption{Ablation of \textbf{feature fusion} on OSD-Amodal}
    \resizebox{\columnwidth}{!}{
        \centering
        \begin{tabular}{|c|c|c|ccccccc|}
        \hline
        \multirow{2}{*}{$F_{\mathcal{B}}$} & \multirow{2}{*}{$F_{\mathcal{V}}, F_{\mathcal{A}}, F_{\mathcal{O}}$} & \multirow{2}{*}{\begin{tabular}[c|]{@{}c@{}}Overall\\ Score\end{tabular}} &
        \multicolumn{2}{c}{Amodal Mask} & \multicolumn{2}{c}{Invisible Mask} & Occlusion & \multicolumn{2}{c|}{Visible Mask} \\
         & &  & OV & $F@.75$ & OV & $F@.75$ & $ACC_{\mathcal{O}}$ & OV & $F@.75$ \\ \hline
                &       & 0.23 & 82.0 & 84.2 & 51.3 & 44.1 & 87.4 & 83.8 & 76.8 \\
        \Checkmark &    & 0.51 & 80.6 & 84.7 & 53.0 & 46.8 & 86.8 & 84.7 & 79.1 \\
           & \Checkmark & 0.56 & 81.2 & 84.4 & 53.0 & 48.2 & 89.5 & 84.3 & 78.0 \\
        \Checkmark & \Checkmark & \underline{\textbf{0.86}} & 82.1 & 83.7 & 55.3 & 49.2 & 90.6 & 85.2 & 79.2 \\ \hline
        \end{tabular}}
        \label{table5}
    \end{table}

    \noindent\textbf{Ablation Studies.} Table \ref{table4} shows the ablation of hierarchy orders in UOAIS-Net on OSD-Amodal with RGB-D. The overall score is the average of the min-max normalized values of each column. Consistent with our motivation, the model with the HOM scheme ($\mathcal{V} \rightarrow \mathcal{A} \rightarrow \mathcal{O}$) performed best, while the reverse scheme ($\mathcal{O} \rightarrow \mathcal{A} \rightarrow \mathcal{V}$) led to the worst performances. This indicates that the proposed HOM scheme is effective in modeling the occlusion of objects. Table \ref{table5} shows the ablation of feature fusion with $F_{\mathcal{B}}$ and $F_{\mathcal{V}}, F_{\mathcal{A}}, F_{\mathcal{O}}$ in the HOM Head of UOAIS-Net on OSD-Amodal. The model with feature fusion of both $F_{\mathcal{B}}$ and $F_{\mathcal{V}}, F_{\mathcal{A}}, F_{\mathcal{O}}$ achieved the best performance, highlighting the importance of dense feature fusion in the prediction heads.
    
\begin{figure}[htb]
    \centering
        \begin{subfigure}[t]{\columnwidth}
            \centering
            \includegraphics[width=\textwidth]{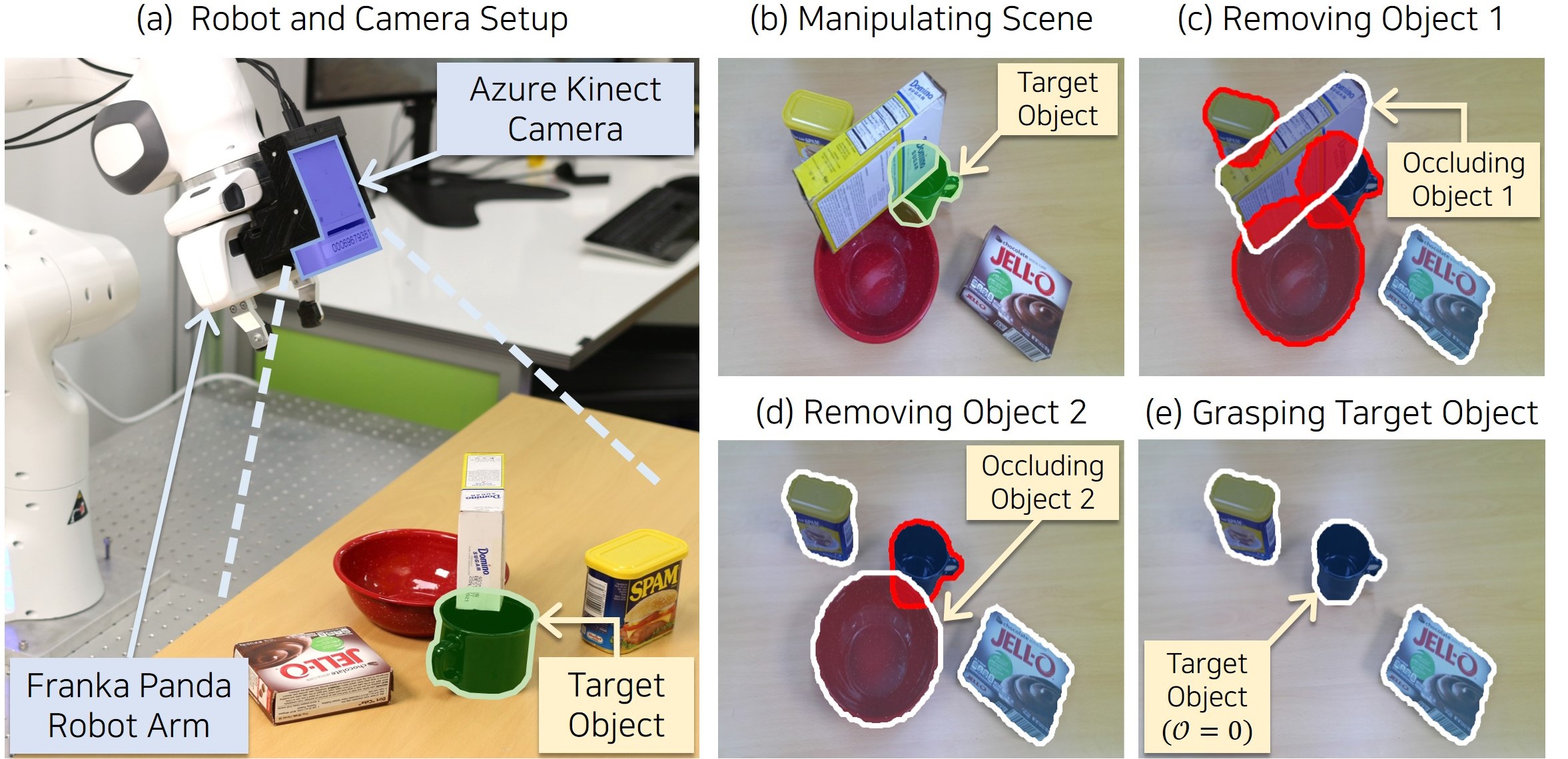}
        \end{subfigure}
    \caption{\textbf{Occluded target retrieval using UOAIS}. To pick up the target object (cup in (a)) in a cluttered scene (b), grasp the unoccluded objects sequentially (box in (c) and bowl in (d)), then the target object can be easily retrieved (e).}
    \label{robotdemo}
    \end{figure}

    \noindent\textbf{Occluded Object Retrieval.} We demonstrated an occlusion-aware target object retrieval with UOAIS-Net (Fig. \ref{robotdemo}). When the target object is occluded, grasping it directly is often infeasible due to the collision between the objects and the robot. UOIS \cite{xiang2020learning, xie2021unseen} can be used to segment the objects but lacks the ability to identify the occlusion relationships, requiring an extra collision checking model \cite{murali20206}. Using UOAIS-Net, the grasping order to retrieve the target object can be easily determined; If the target object is occluded ($O=1$), remove the nearest unoccluded objects ($O=0$) until the target becomes unoccluded ($O=0$). Then the target object can be easily grasped. We used Contact-GraspNet \cite{sundermeyer2021contact} for grasping and a ResNet-34 \cite{he2016deep} to classify the objects.
    
\begin{figure}[ht!]
        \centering
        \includegraphics[width=\columnwidth]{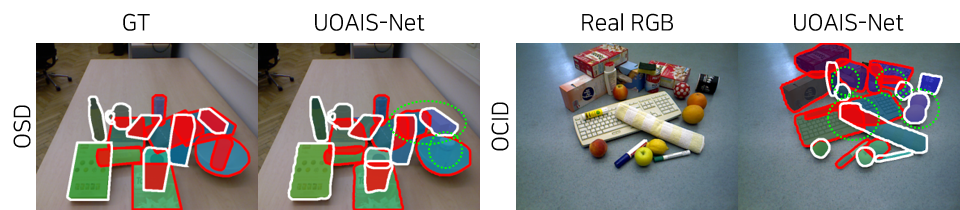}
        \caption{Failure Cases of UOAIS-Net in OSD and OCID}
        \label{fig:failure}
    \end{figure}

    \noindent\textbf{Failure Cases} Fig. \ref{fig:failure} shows some failure cases of UOAIS-Net with RGB-D. When the object is on the other object, it sometimes over-segments the occluded objects. As a result, it degraded the UOAIS-Net performance in amodal masks on OSD, though the performance of our model in visible and invisible masks is noticeably better than others. We believe this could be resolved by increasing the quality and quantity of the synthetic dataset and leaving it as future work.

\section{Conclusion}
This paper proposed a novel task, UOAIS, to jointly detect the visible masks, amodal masks, and occlusions of unseen objects in a single framework. Under the HOM scheme, UOAIS-Net could successfully detect and reason the occlusion of unseen objects with SOTA performance on the three datasets. We also demonstrated a useful demo for occluded object retrieval. We hope UOAIS can serve as a simple and effective baseline for amodal robotic manipulation.

\section{Acknowledgement}
\begin{spacing}{0.4}
{\scriptsize We give a special thanks to Sungho Shin and Yeonguk Yu for their helpful comment and technical advice. This work was fully supported by the Korea Institute for Advancement of Technology (KIAT) grant funded by the Korea Government (MOTIE) (Project Number: 20008613). This work used computing resources from the HPC Support project of the Korea Ministry of Science and ICT and NIPA.}
\end{spacing}





\bibliography{bibtex/bib/references.bib}{}
\bibliographystyle{IEEEtran}

\end{document}